\documentclass[article, 9pt]{IEEEtran}

\usepackage{amsmath,amssymb,graphicx}
\usepackage{subfig}
\usepackage{epsfig}
\usepackage{epstopdf}
\usepackage{float, cite}
\usepackage{amsfonts,amsthm}		
\usepackage{color}
\usepackage{caption}
\usepackage{algorithm}
\usepackage{algorithmic}

\def \defn{\stackrel{\Delta}{=}}

\def \ie{{i.e.}}
\def \vx{{\mathbf x}}

\def \vw{{\mathbf w}}
\def \vbeta{{\mathbf \beta}}
\def \vD{{\mathbf D}}
\def \vM{{\mathbf M}}

\def \v0{{\mathbf 0}}
\def \vC{{\mathbf C}}
\def \vSig{{\mathbf \Sigma}}

\def \vc{{\mathbf c}}

\def \vy{{\mathbf y}}
\def \vY{{\vec Y}}

\def \vU{{\mathbf U}}
\def \vLambda{{\mathbf \Lambda}}
\def \vS{{\mathbf S}}

\def \vA{{\mathbf A}}
\def \E{\mathbb E}
\def \vI{{\mathbf I}}
\def \tr{{\mathrm{tr}}}
\def \vSigma{{\boldsymbol{\Sigma}}}

\def \vx{{\mathbf x}}
\def \vc{{\mathbf c}}

\def \vy{{\mathbf y}}

\def \v0{{\mathbf 0}}
\def \vI{{\mathbf I}}
\def \vA{{\mathbf A}}
\def \vB{{\mathbf B}}

\def \vM{{\mathbf M}}

\def \vS{{\mathbf S}}
\def \vV{{\mathbf V}}
\def \vW{{\mathbf W}}
\def \vQ{{\mathbf Q}}

\def \vY{{\mathbf Y}}

\def \bbR{{\mathbb R}}

\def \tr{{\text{tr} }}

\def \Diag{{ \textbf{Diag} }}

\IEEEoverridecommandlockouts

\title{Compressive Measurement Designs for Estimating\\
Structured Signals in Structured Clutter:\\
A Bayesian Experimental Design Approach\thanks{The authors are with the Dept. of Electrical and Computer Engineering, University of Minnesota, Minneapolis, MN; email: {\tt \{jainx174, sonix022, jdhaupt\}@umn.edu}. This work was supported by DARPA/ONR Award No. N66001-11-1-4090.}}

\author{Swayambhoo Jain, Akshay Soni, and Jarvis Haupt}

\begin{document}

\maketitle

\begin{abstract}
This work considers an estimation task in compressive sensing, where the goal is to estimate an unknown signal from compressive measurements that are corrupted by additive pre-measurement noise (interference, or ``clutter'') as well as post-measurement noise, in the specific setting where some (perhaps limited) prior knowledge on the signal, interference, and noise is available.  The specific aim here is to devise a strategy for incorporating this prior information into the design of an appropriate compressive measurement strategy.  Here, the prior information is interpreted as statistics of a prior distribution on the relevant quantities, and an approach based on Bayesian Experimental Design is proposed.  Experimental results on synthetic data demonstrate that the proposed approach outperforms traditional random compressive measurement designs, which are agnostic to the prior information, as well as several other knowledge-enhanced sensing matrix designs based on more heuristic notions.
\end{abstract}


 \section{Introduction}

This paper investigates a problem in ``sensing matrix'' design arising in the context of certain compressive sensing (CS) estimation tasks.  Let $\vx \in \mathbb{R}^n$ represent the object we aim to estimate, and suppose that we obtain $m$ noisy measurements of $\vx$ as follows
\begin{equation}\label{eqn:measure}
\vy = \vA (\vx+\vc) + \vw,
\end{equation}
where $\vA$ is the $m\times n$ sensing matrix, $\vc$ as a $n\times 1$ vector of pre-measurement interference or ``clutter,'' and 
$\vw\in\mathbb{R}^m$ is a vector of perturbations whose elements may describe additive measurement noise or modeling error.  
Investigations of problems of this form in the so-called underdetermined setting ($m<n$) have been the primary focus of recent efforts in 
CS.  Indeed, a primary focus in CS research has been the analysis of sensing and inference procedures for estimating $\vx$ from such noisy 
linear measurements in the case where $\vx$ is \emph{sparse}, having, say, $k < n$ nonzero or significant entries. 

Initial efforts in CS analyzed the task of estimating $\vx$ from observations obtained according to \eqref{eqn:measure}, in the case where there is no clutter ($\vc=\v0$) and the additive noise vector $\vw$ is assumed to be $\mathcal{N}(0,\sigma^2\vI_{n})$ distributed. In these settings it is now known that only $m=O(k \log n)$ compressive measurements suffice to obtain an estimate $\widehat{\vx}$ satisfying $\|\vx - \widehat{\vx}\|_2^2 \leq \mbox{const.} \ k \sigma^2 \log n$ with high probability (see, for example, \cite{Dantzig}).  Several recent works have examined the effects of clutter (\ie, the case $\vc \neq \v0$) in compressive sensing estimation tasks, but these investigations have typically been limited to the case where the clutter is modeled as white Gaussian noise \cite{Reeves, Saligrama, Arias}. One exception is the work \cite{Krishnamurthy} which, in the context of a different compressive inference task (target detection), utilizes an observation model analogous to \eqref{eqn:measure} in which the clutter is assumed to be a realization of a Gaussian random vector having nonzero mean and non-trivial covariance matrix.  Ultimately, the approach employed in that work is to view the image of the clutter under $\vA$ as another noise contribution, and to compensate for its effect by ``whitening'' the compressive measurements $\vy$. 

A notable aspect of the result \cite{Dantzig} (in the clutter-free scenario) and indeed, many related results in the CS literature, is that \emph{random} matrices $\vA$ whose elements are drawn iid from certain zero-mean distributions comprise a broad class of sensing matrices that facilitate accurate estimation of sparse $\vx$ in CS (see, for example, \cite{Baraniuk}). The ``universality'' of such approaches is often cited as virtue, since the same $\vA$ can suffice for acquiring \emph{any} sparse (enough) $\vx$.  On the other hand, in many scenarios we may be equipped with additional information about the signal we aim to estimate, beyond simply an assumption of sparsity (e.g., that the signal possess a structured sparse representation).  This additional information can, of course, be incorporated into the inference task to improve estimation performance \cite{Huang, ModelBased}.  On the other hand, a unique (in fact, \emph{essential}) assumption underlying the CS paradigm is the ability to obtain \emph{generalized} linear measurements of the quantity of interest.  This inherent flexibility of the measurement process suggests that we should consider incorporating (in a principled manner, and as appropriate) the additional information directly into the design of the \emph{sensing} process.

Here we focus on a \emph{knowledge-enhanced} estimation problem associated with the compressive measurements obtained via the model \eqref{eqn:measure}.  Our aim remains to estimate $\vx$, and we assume that we are equipped with some additional \emph{prior knowledge} about $\vx$, $\vc$, and $\vw$.  The prior knowledge about $\vx$ could describe, for example, a small collection of possible supports (locations where $\vx$ takes its nonzero values) and their relative frequencies of occurrence, or correlation structure among the nonzeros of $\vx$.  Likewise, prior knowledge about $\vc$ and $\vw$ may identify the correlation structures or supports of each.  The question we address here is, how should we design the sensing matrix $\vA$ to take advantage of this prior knowledge?

The main contribution of this work is to demonstrate that knowledge-enhanced sensing matrix designs can outperform purely random sensing matrix designs in noisy compressive sensing tasks in which the goal is to \emph{estimate structured signals in structured clutter}.  Here, we describe the available prior information on the quantities $\vx$, $\vc$, and $\vw$ in terms of distributions with known first- and second-order statistics. Using ideas from Bayesian Experimental Design \cite{chaloner}, we formulate the sensing matrix design task as an optimization whose aim is to minimize the mean-square error (MSE) of associated with the \emph{best}\footnote{As we will describe in the next section, here we restrict the class of estimators to be \emph{linear} in the observations $\vy$, for analytical tractability.} estimator of $\vx$ obtained from the measurements $\vy$.  We state a simple procedure for obtaining the solution to corresponding optimization task, and demonstrate via simulation the performance improvements resulting from our approach relative to random CS measurement designs.  Our experimental analysis also includes comparison with several other knowledge-enhanced measurement designs based on more heuristic notions.

The remainder of this paper is organized as follows.  Following a brief discussion of our contribution in the context of existing work (below), we formally describe our problem in Section~\ref{sec:setting}.  Our main result -- a simple algorithmic approach for knowledge-enhanced compressive sensing matrix design -- is presented in Section~\ref{sec:KECoM}. We provide experimental validation of our approach in Section~\ref{sec:eval}, and briefly discuss extensions in Section~\ref{sec:conc}.  

\begin{table*}[t]
\centering \vspace{-3mm}
\begin{tabular}{|l|}
\hline
Successive minimization algorithm to solve :  $\underset{\vA \in \bbR^{m \times n}}{ \text {maximize}} \quad \tr\left( \vSig_x  \vA 
\left(  \vA \left( \vSig_x + \vSig_c \right) \vA' + \vI  \right)^{-1}  \vA'  \vSig_x' \right) \text{ subject to}  ~\| \vA\|_F^2 \le 
\alpha^2$ \\
\hline
Input: Covariance matrices $\vSig_x$ and $\vSig_c$, budget parameter $\alpha$, number of iterations $N$ \\
1: Find $\vY$ such that $\vY' (\vSig_x + \vSig_c) \vY = \vI_m$. \\
2: Calculate eigendecomposition  of $\vY' \vSig_x^2 \vY$, denoted $\vU_{1} \vLambda_{1} \vU_{1}'=\vY' \vSig_x^2 \vY$ \\
3: Initialize $\{\sigma^0_i\}_{i=1}^m$ \\
Repeat: 4 and 5 for $j=1$ to $N$ \\
\quad 4:  Update $\vU_M^j$: \\ 
 \quad  \quad \qquad  (i) Form $\vU_M^j$ by $m$ columns of $\vU_{1}$ such that $  \tr\left(  \vU_M' \vY' \vSig_x^2 \vY \vU_M \Diag\left(
 \frac{(\sigma_1^j)^2}{1+(\sigma_1^j)^2}, \cdots, \frac{(\sigma_m^j)^2}{1+(\sigma_m^j)^2}
 \right) \right)$ is maximized. \\ 
\quad 5:  Update $\sigma^j_i$ \\
 \quad  \quad \qquad  (i) Compute $b_{i}$ and $c_{i}$ as the $i^{th}$ diagonal entries of $(\vU_M^{j-1})' \vY' \vSig_x^2 \vY  (\vU_M^{j-1})$ and $(\vU_M^{j-1})' \vY' \vY (\vU_M^{j-1})$, respectively.\\
\quad  \quad \qquad  (ii) Solve: $\sigma_i^2 = \left( \sqrt{\frac{b_i}{c_i v}} -1 \right)^+$ and  $\sum_{i=1}^m c_i \left( 
\sqrt{\frac{b_i}{c_i v}} -1 \right)^+ = \alpha^2$ (via waterfilling-based approach)\\
6: Compute $\vM = \vU_M^{N} \Diag\left(
\sigma_1^N,\cdots, \sigma_n^N\right) $ \\
Output: $\vA =  (\vY\vM)' $\\
\hline
\end{tabular}
\caption{Iterative algorithm for solving the sensing matrix design problem \eqref{eqn:opt2}.}
\label{tab:alg1}
\vspace{-8mm} 
\end{table*}

\subsection{Connections with Prior Works}

The work \cite{Elad} proposed one of the first approaches to design compressive sensing matrices given some prior signal knowledge.  That work considered noise-free settings and assumed knowledge of a dictionary in which the signals being observed were sparse, and proposed a sensing matrix design procedure whose aim is to reduce the coherence between the learned sensing matrix and the known dictionary.  Extensions of this idea aimed at designing both the dictionary and the sensing matrix given a collection of training data were examined by \cite{Sapiro1} for the case of ``simple'' sparsity, and \cite{EldarDesign} for signals possessing a block-sparse representation in a known dictionary.  

The recent work \cite{Sapiro2} examined knowledge-enhanced CS design tasks using a probabilistic formulation of the prior knowledge, as here.  That work assumed a Gaussian mixture prior on the signal being acquired, and proposed a design criteria based on coherence minimization between the learned sensing matrix and a dictionary composed of eigenvectors of the mixture covariance matrices.  Along the same lines, the work \cite{Rao} examined sensing designs based on learned correlations in training data.  We note that none of these approaches utilize the statistical estimation theoretic formulation we adopt here. Our effort here is also related to the body of prior work on optimal designs for space-time linear coding in MIMO applications -- see, for example, \cite{Scaglione}, which examined qualitatively similar estimation problems but without the additive interference or ``clutter'' term. 

Our effort is also related to existing works that examined Bayesian experimental design problems in compressive sensing estimation tasks \cite{BCS, seeger, seeger2, seeger3, schniter}, and subsequent efforts along these lines examined the performance improvements resulting from Bayesian experimental design strategies in some specific application domains (e.g., magnetic resonance imaging applications \cite{seeger_mri}).  These efforts are utilize a design principle based on maximizing the mutual information between the vector $\vx$ to be estimated, and the observations $\vy$ obtained with the designed matrix $\vA$.  Similar mutual information maximization criteria were utilized in the recent work \cite{carson}, which considered a Bayesian analog of the sensing matrix design task first proposed in \cite{Elad}.

It is worth noting that none of the aforementioned works deal explicitly with separation from clutter, in the case where we have essentially no interest in \emph{estimating} the clutter $\vc$.  Rather, in our formulation, its presence is more akin to a nuisance parameter in our overall estimation task.  In this sense, our problem is related also to the wealth of classical work on \emph{interference cancellation} (see, for example \cite{van trees}), but with a compressive sensing/sparse inference ``twist.''

\section{Problem Statement}\label{sec:setting}

As alluded above our ultimate inference goal is to accurately estimate the vector $\vx$ given measurements obtained according to \eqref{eqn:measure}, in settings where we may design the sensing matrix $\vA$ using prior information about the signal, clutter, and noise.  In this section we describe our overall sensing matrix design methodology.

\subsection{Quantifying Prior Information}

In our approach here we will assume that the vector $\vx\in\mathbb{R}^n$ that we wish to estimate is a random quantity drawn from a mixture distribution having $m_x$ mixture components.  We do not assume full knowledge of the mixture distribution, but only that $i$-th mixture component has known weight $\pi_{x,i}$ and is an $n$-dimensional zero-mean random vector with known $n\times n$ covariance matrix $\vSig_{x,i}$, for $i=1,2,\dots,m_x$. 

We note that the covariance matrices $\vSig_{x,i}$ are not assumed here to be full-rank.  On the contrary, rank-deficiency in any of the $\vSig_{x,i}$ amounts to a form of \emph{sparsity}, as random vectors $\vx\in\mathbb{R}^n$ drawn from a distribution with covariance matrix of rank $r<n$ inherently lie on a $r$-dimensional subspace of $\mathbb{R}^n$.  Thus, the formulation described here can model various forms of sparsity and structure that have been studied in the literature, including simple $k$-sparse vectors (the collection of $\vSig_{x,i}$ describe all $m_x = {n \choose k}$ unique subsets of $\{1,2,\dots,n\}$ of cardinality $k$, and $\pi_i = 1/m_x$), block sparsity, group sparsity (with potentially overlapping groups), tree sparsity, and so on.  
It is worth noting that our model does not assume that vectors drawn from different models be orthogonal, though that structure, if present, could easily be captured by this formulation.  

Likewise, we assign an analogous prior distribution to the clutter $\vc$, modeling it as a realization of an $m_c$-component mixture distribution whose $i$-th mixture component has weight $\pi_{c,i}$ and is a zero-mean random vector with covariance matrix $\vSig_{c,i}$, for $i=1,2,\dots,m_c$. We consider $\vw$ to be additive uncorrelated zero-mean noises with unit variance, and we assume that the random quantities $\vx$, $\vc$, and $\vw$ are uncorrelated. 

\subsection{Minimizing the Estimation MSE}

Our aim here is to minimize the mean-square error (MSE)
associated with our ultimate estimate of the signal $\vx$.  Formally, we denote by $\widehat{\vx}_{\vA}(\vy)$ an \emph{estimate} of $\vx$ obtained using a particular estimation \emph{strategy}, denoted here by $\widehat{\vx}_{\vA}$.  Note that the estimation strategy is parameterized by the sensing matrix $\vA$, and a particular estimate obtained using this strategy is a function of the measurements $\vy$ obtained via \eqref{eqn:measure} using that $\vA$.  The mean-square error associated with a particular estimation strategy $\widehat{\vx}_{\vA}$ is denoted by $d_{\rm MSE}(\widehat{\vx}_{\vA})\defn  \E_{\vx,\vc,\vw}\left[\|\vx - \widehat{\vx}_{\vA}(\vy)\|^2\right]$, where the subscript denotes that the expectation is with respect to all of the random quantities.   The criteria for optimal design of the sensing matrix $\vA$ in this case can be stated as an optimization -- the optimal choice of $\vA$, denoted by $\vA^*$, is 
\begin{equation}\label{eqn:design}
\vA^* = \arg \min_{\vA \in \mathcal{A}} \ \min_{\widehat{\vx}_{\vA} \in \mathcal{X}} \ d_{\rm MSE}(\widehat{\vx}_{\vA}),
\end{equation}
where $\mathcal{A}$ is a (possibly constrained) class of sensing matrices and $\mathcal{X}$ is a (possibly constrained) class of possible estimation strategies.  In words, $\vA^*\in\mathcal{A}$ is the sensing matrix yielding measurements for which the MSE of the \emph{best possible} estimation strategy (from the class $\mathcal{X}$) is minimum.    

\sloppypar Note that the presence of the measurement noise $\vw$ is only relevant when the sensing matrix $\vA$ is constrained in some way.  Indeed, in unconstrained settings simply scaling each of the elements of $\vA$ toward infinity would make the overall effect on $\vw$ negligible in the estimation task. Here our focus will be on \emph{energy-constrained} designs $\vA$; in particular, we choose $\mathcal{A}$ in \eqref{eqn:design} as $\mathcal{A} = \{\vA: \|\vA\|_F \leq \alpha\}$ for some (specified) $\alpha>0$,
where the notation $\|\cdot\|_F$ denotes the matrix Frobenius norm.  Each row of $\vA$ is itself a linear operator which gives rise to one (noisy, cluttered) compressive sample; thus, the constraint we impose here amounts to a constraint on the average energy per-row in the sensing matrix.

\section{Sensing Matrix Designs for Minimum MSE}\label{sec:KECoM}
It is well-known from statistical estimation theory that, for the minimum MSE task (MMSE) task described above, the optimal estimator of $\vx$ is the conditional mean $\vx$ given the observations $\vy$; that is, $\widehat{\vx}_{\vA,{\rm MMSE}}(\vy) = \E\left[\vx | \vy\right]$ (see, for example, \cite{kay1}). Here, our prior knowledge is limited to first- and second-order statistics of the signal, clutter, and noise, and without full knowledge of the distributions we are unable to compute this estimator in closed form.  Instead, we consider restricting the class of estimators $\mathcal{X}$ in \eqref{eqn:design} to be the class of \emph{linear} estimators of $\vx$, as described below.

We define the \emph{average} signal covariance matrix $\vSig_x$ as $\vSig_x = \sum_{i=1}^{m_x} \pi_{x,i} \vSig_{x,i}$,
and similarly for $\vSig_c$, and we assume that $(\vSig_x + \vSig_c)$ is invertible. Now, the linear MMSE estimator is just the Wiener Filter, easily shown here to be $\widehat{\vx}_{\vA,{\rm LMMSE}}(\vy) = \vSig_x \vA'\left(\vA\left(\vSig_x + \vSig_c\right)\vA' + \vI_{n}\right)^{-1}\vy$,
where $\vA'$ denotes the matrix transpose. It follows (after a bit of algebra) that 
\begin{eqnarray}
\lefteqn{\E_{\vx,\vc,\vw}\left[\|\vx - \widehat{\vx}_{\vA,{\rm LMMSE}}(\vy)\|^2\right] =}&&\\
\nonumber && \tr \{\vSig_x - \vSig_x\vA'\left(\vA\left(\vSig_x + \vSig_c\right)\vA' + \vI_{m}\right)^{-1}\vA\vSig_x \},
\end{eqnarray}
where $\tr\{\cdot\}$ denotes the matrix trace (the sum of the diagonal elements). Thus, we can express our sensing matrix design task as an optimization, whose aim is to minimize the trace of the estimation error covariance matrix\footnote{In the parlance of Bayesian experimental design, this corresponds to a simple instance of a Bayes $A$-optimality criteria.}.  Here\footnote{A similar problem was addressed in \cite{schizas2007distributed}, but under a \emph{transmit energy} constraint of the form $\tr\left(\vA \left(\vSig_x + \vSig_c\right)\vA^T\right) \le \alpha^2$. That said, the solution approach therein seems to be specific to the \emph{transmit energy} constraint, and can not be directly extended to address the \emph{sensing energy} constraint $\tr\left(\vA \vA^T\right) \le \alpha^2$ we impose here.}, this amounts to an optimization problem
\begin{eqnarray}\label{eqn:opt2}
\lefteqn{\vA^* =}&&\\
\nonumber &&\hspace{-2em} \arg \max_{\vA: \|\vA\|_F \leq \alpha} \ \tr\left\{\vSig_x \vA'\left(\vA\left(\vSig_x + \vSig_c\right)\vA' +  \vI_{m}\right)^{-1}\vA\vSig_x\right\}.
\end{eqnarray}
Our preliminary investigation on this problem (reported in \cite{SPARS}) entailed a solution approach for \eqref{eqn:opt2} that utilized an approximation of the inverse term in the objective, and led to a design strategy whose applicability was valid only in qualitatively low-SNR regimes.  In the following subsection we describe an approach for obtaining the solution to \eqref{eqn:opt2} for in general settings, and for various sizes of sensing matrices $\vA$. 
\begin{figure*}[t]
\centering
\vspace{-8mm} 
$\begin{array}{cccc}
\includegraphics[scale=0.23]{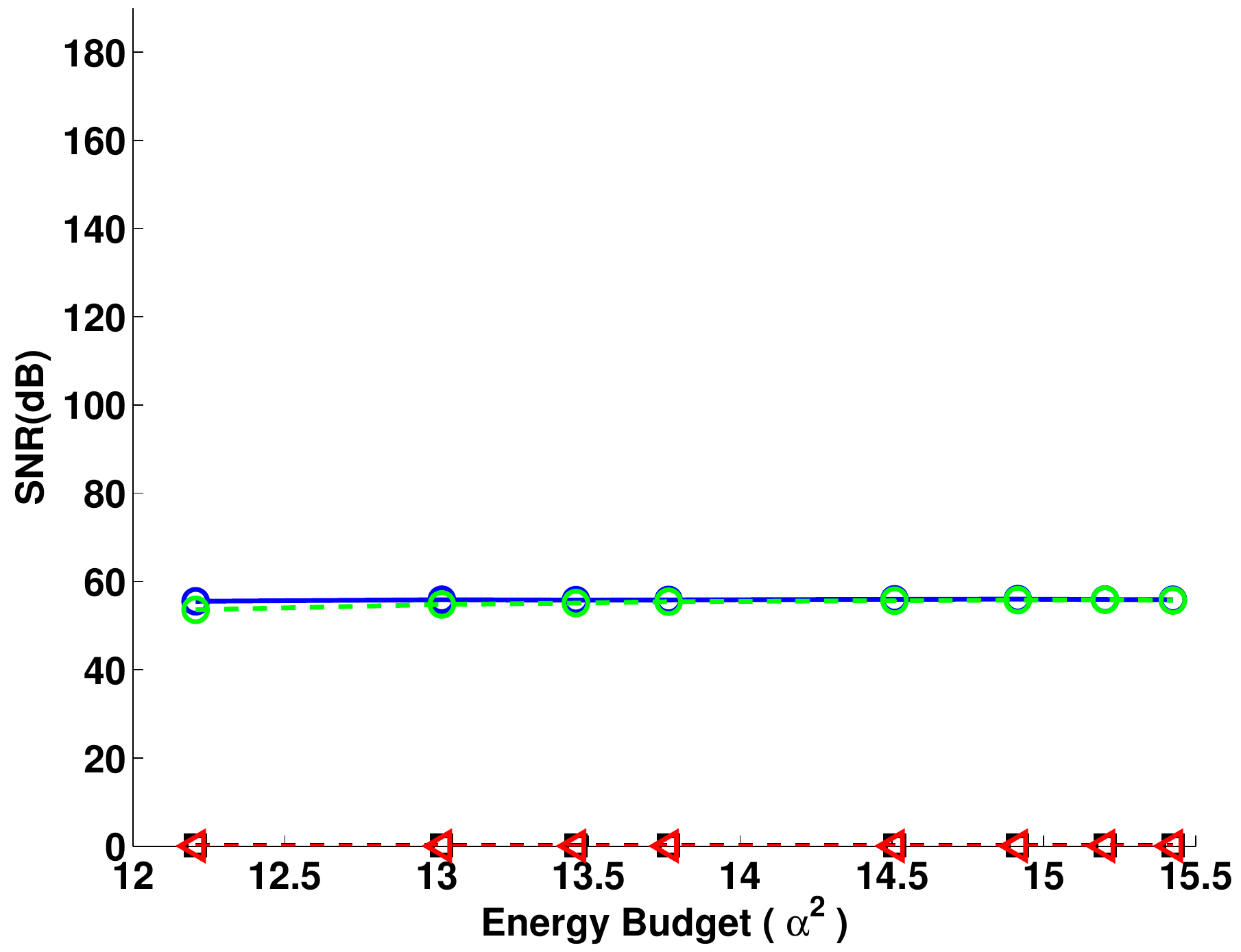} & \includegraphics[scale=0.23]{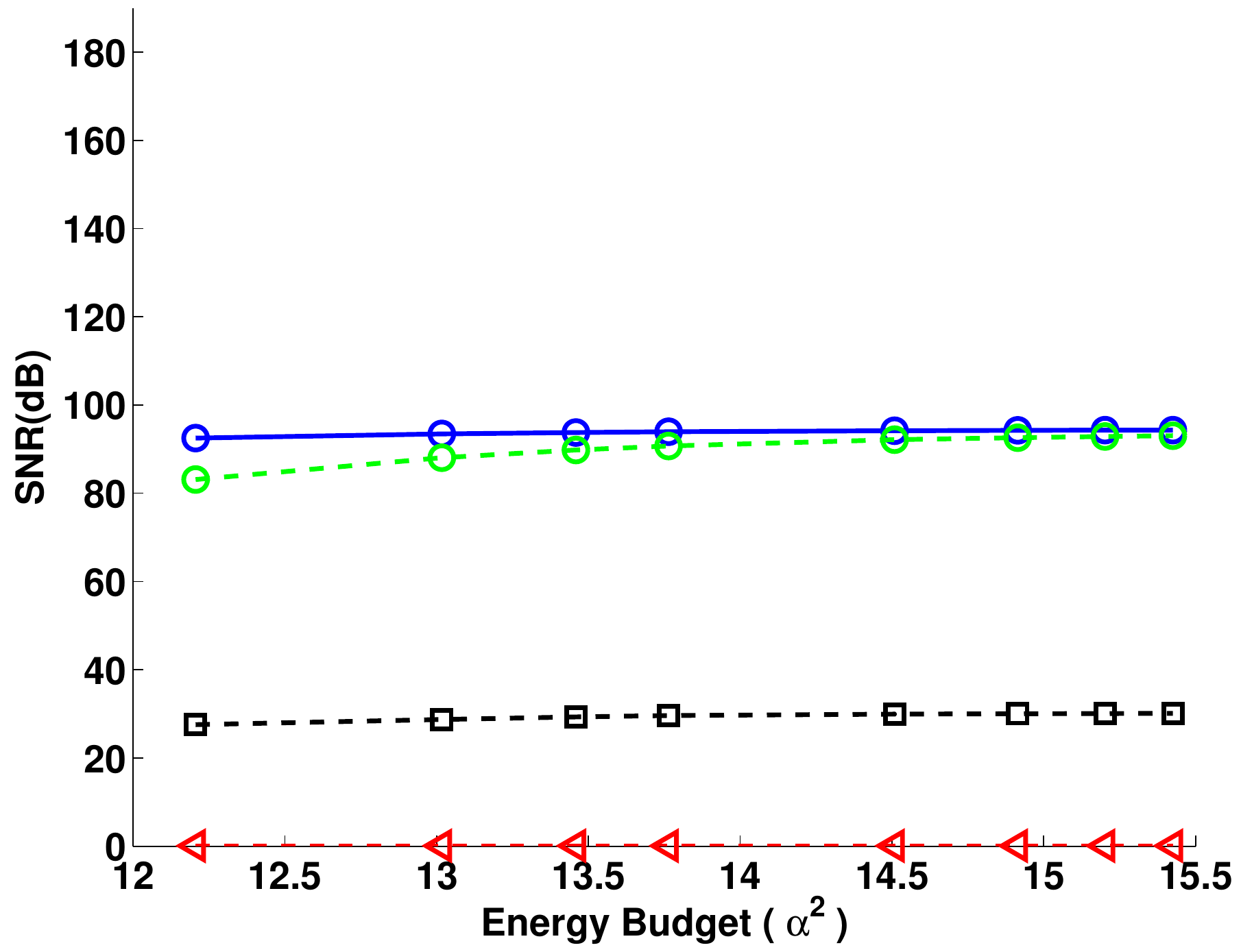} &
 \includegraphics[scale=0.23]{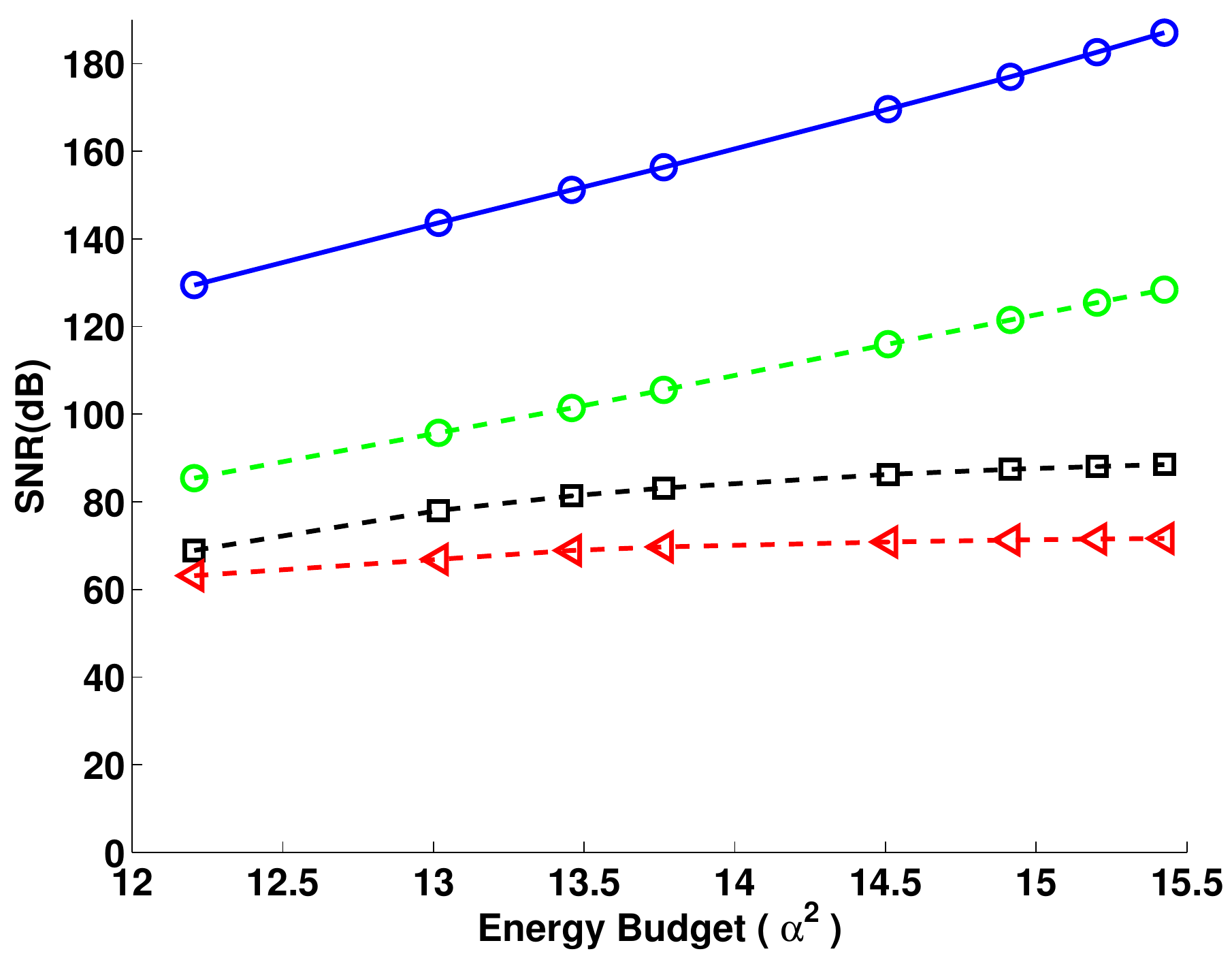} & \includegraphics[scale=0.23]{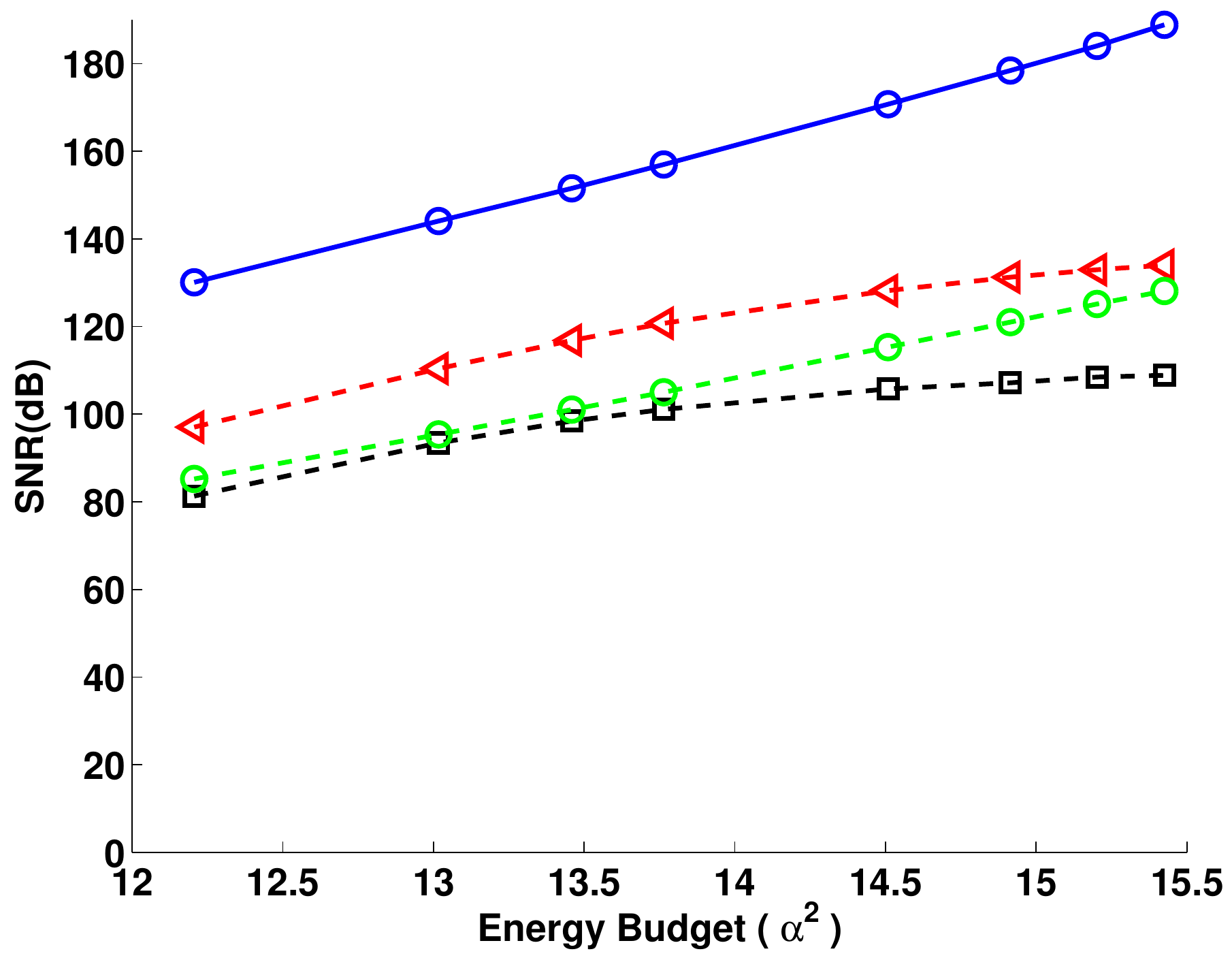}\\
(a) & (b)&(c) & (d)
\end{array}$
	\caption{Reconstruction SNR = $20 \log \frac{\|\vx\|_2}{\|\vx - \hat{\vx}\|_2}$ vs. sensing energy budget for several different compressive measurement strategies (see text for details).  Panels (a)-(d) correspond to $m=20, 40, 60, 80$ measurements, respectively.  Higher SNR values correspond to better reconstructions.  Our proposed approach (blue dotted line, circle markers) outperforms each of the other measurement strategies examined. }
	\label{fig:res}
	\vspace{-8mm} 
\end{figure*}
\vspace{-3mm} 
\subsection{Our Proposed Design Approach}
For solving \eqref{eqn:opt2} we make the following variable transformation: let $\vA' = \vY \vM$ , where $\vY$ is $n \times n$  full rank matrix satisfying\footnote{Since $\vSig_x + \vSig_c$ is positive definite, we can always find a diagonalizing matrix $\vY$ from the eigenvalue decomposition of $\vSig_x + \vSig_c$. Specifically, let $\vSig_x + \vSig_c = \vU_{\vx + \vc } \vSigma_{\vx + \vc} \vU_{\vx + \vc } '$, then $\vY = \vU_{\vx + \vc }  \Sigma_{\vx + \vc}^{-1/2}.$ 
Overall there can be many choices of $\vY$ which diagonalize $\vSig_x+\vSig_c$; in fact, $\vSig_x + \vSig_c$ is diagonalized by $\vY$ then it is also diagonalized by $\vY \vQ$ for any orthonormal matrix $\vQ$.}
 \begin{align}
 \vY'(\vSig_x + \vSig_c  )\vY  = \vI_n,
 \end{align}
 and $\vM $ any $n \times m$ matrix.  Further,  using the thin singular value decomposition of $\vM= \vU_M \vSigma_M \vV_M'$, where $\vU_M \in \bbR^{n \times m}$ with $\vU_M' \vU_M = \vI_m $, $\vSigma_M  = \Diag(\sigma_1,\cdots,\sigma_m)$  with $\sigma_i \ge 0 \quad \forall i = 1 \dots m$, and $\vV_M \in \bbR^{m\times m}$ is an orthonormal matrix, we can recast the problem \eqref{eqn:opt2} after a bit of linear algebra as
\begin{align}
\begin{array}{cl}
\underset{\stackrel{\vU_M \in \bbR^{n \times m}}{\vSigma_M \in \bbR^{m \times m}}}{ \text {maximize}} & \tr\left( \vSigma_M' \vU_M' 
\vY'\vSig_x^2 \vY \vU_M \vSigma_M \left(   \vSigma_M^2+ \vI_m  \right)^{-1}  \right) \\ 
\text{subject to } &  \vU_M' \vU_M = \vI_m,  ~ \tr(\vSigma_M \vY'\vY   \vU_M \vSigma_M) \le \alpha^2 \\ 
& \vSigma _M= \Diag(\sigma_1,\cdots,\sigma_m) \succeq \v0, 
\label{prob: P4.5} 
\end{array}
\end{align}
where the notation $\vSigma _M \succeq \v0$ denotes that $\vSigma _M$ is positive semidefinite.
Now since $\vSigma_{M}$ is diagonal so we can equivalently write the above problem as 
\begin{align}
\begin{array}{cl}
\underset{\stackrel{\vU_M \in \bbR^{n \times m}}{\sigma_i \ge 0 }}{ \text {maximize}} & \tr\left(  \vU_M' \vY'\vSig_x^2 \vY \vU_M\tilde{\vSig}  \right) \\ 
\text{subject to } &  \vU_M' \vU_M = \vI_m,  ~  \tr( \vU_M' \vY' \vY  \vU_M \vSigma_M^2) \le \alpha^2,
\end{array}
\label{prob: P5} 
\end{align}
where $\tilde{\vSig} = \Diag\left(
\frac{\sigma_1^2}{1+\sigma_1^2}, \cdots, \frac{\sigma_m^2}{1+\sigma_m^2}
\right) $. The problem \eqref{prob: P5} is a non-convex problem, so we resort to successive minimization over $\{\sigma_i\}_{i=1}^m$  and $\vU_M$ by successively solving the following subproblems 
\begin{align}
\begin{array}{lll}
\mathbf{P_0} : & \underset{\{\sigma_i\}_{i=1}^m; \ \sigma_i \ge 0 \ \forall i}{ \text {maximize}} & \tr\left(  \vU_M' \vY'\vSig_x^2 \vY \vU_M \tilde{\vSig} \right) \\ 
& \text{subject to } &  \tr(\vSigma_M \vU_M' \vY' \vY  \vU_M \vSigma_M) \le \alpha^2. \\
\mathbf{P_1}:  & \underset{\vU_M \in \bbR^{n \times m}}{ \text {maximize}} & \tr\left(  \vU_M' \vY'\vSig_x^2 \vY \vU_M \tilde{\vSig} \right) \\ 
&\text{subject to } &  \vU_M' \vU_M = \vI_m. \\ 
\end{array} 
\end{align}

The sub-problem $\mathbf{P_0}$ is maximization over $\{\sigma_i\}_{i=1}^m$ for fixed $\vU_M$, and $\mathbf{P_1}$ is maximization over $\vU_M$ for fixed $\{\sigma_i\}_{i=1}^m$. The main novelty here is in the way we split the constraints. In the following subsections we demonstrate how to solve these sub-problems.

\subsubsection{Solving \textbf{P0}}\label{subsection: Pa}
With some linear algebra we can show \textbf{P0} is equivalent to 
\begin{align}
\begin{array}{cl}
\underset{\gamma_i \in \bbR \forall i=1\dots m}{ \text {maximize}} & \sum_{i=1}^m \frac{b_{i}\gamma_i}{ 1+ \gamma_i} \\ 
 \text{subject to } &  \sum_{i=1}^m c_{i} \gamma_i \le \alpha^2 \\ 
 & \gamma_i \ge 0, \quad  i = 1, \cdots m.
\end{array} 
\end{align}
where $ \gamma_i = \sigma_i^2 $, and $b_{i}$ and $c_{i}$ are the $i^{th}$ diagonal entry of $\vU_M' \vY'\vSig_x^2 \vY \vU_M$ and $\vU_M' \vY' \vY \vU_M$ respectively. Since $\vY'\vSig_x^2 \vY \succeq \v0 $ all $b_i$'s are non-negative, and since the diagonal entries of $\vY' \vY$ are strictly positive all the $c_i$'s are strictly positive. With this, we can show that \textbf{P0} is a convex problem whose solution is given by
\begin{align}
\gamma_i^* = \left( \sqrt{\frac{b_i}{c_i v^*}} -1 \right)^+ \text{ and } \quad \sum_{i=1}^m c_i \left( \sqrt{\frac{b_i}{c_i v^*}} -1 
\right)^+ = \alpha^2,
\end{align}
where $(a)^+ = \mathrm {max} \{0,a\}$ and  $v^*$ is the Lagrangian multiplier associated with the constraint $ \sum_{i=1}^m c_{i} \gamma_i 
\le \alpha^2$ which can be easily obtained by binary search algorithm as done in standard water-filling solution with a minor modification.

\subsubsection{Solving \textbf{P1}} \label{subsection: Pb}

The Lagrangian for problem \textbf{P1} is 
\begin{equation}
\mathcal{L}( \vU_M, \vS) =  \tr\left(\vU_M' 
\vY'\vSig_x^2 \vY \vU_M \tilde{\vSig} \right) + \tr\left(\vS\left(\vU_M'\vU_M - \vI_m\right)\right), \nonumber
\end{equation}
where $\vS$ is an $m \times m$  symmetric Lagrange multiplier matrix. It can be shown that the orthonormality constraint $\vU_M'\vU_M = \vI_m$ satisfies the \emph{regularity} condition \cite{bertsekas1999nonlinear}, which implies that for every local maxima there exists a unique Lagrange multiplier matrix $\vS$. Taking the gradient with respect to $\vU_M$, and equating it to zero we get  the following local optimality condition
\begin{equation}
\vY'\vSig_x^2\vY\vU_M\tilde{\vSig} = \vU_M\vS. \label{local_opt}
\end{equation}
For $\vU_M$ satisfying \eqref{local_opt}, we have that at the optimal function of \textbf{P1} takes the value 
\begin{equation}
\tr\left(\vU_M'\vY'\vSig_x^2\vY\vU_M\tilde{\vSig}\right) = \tr\left(\vU_M'\vU_M\vS\right) = \tr\left(\vS\right). \nonumber
\end{equation}
This implies that the objective function in \textbf{P1} evaluated at these local maxima depends only on the diagonal entries of $\vS$, so without loss of generality we can restrict ourselves to diagonal $\vS$ in order to find the optimal solution of \textbf{P1}, and can instead solve the following problem which is equivalent to \textbf{P1}
\begin{equation}
\begin{array}{ll}
 \underset{\vU_M \in \bbR^{n \times m}, \vS \in \bbR^{m \times m}}{ \text {maximize}} & \tr\left( \vS \right) \\ 
\text{subject to } &  \vU_M' \vU_M = \vI_m, \\   
& \vY'\vSig_x^2\vY\vU_M\tilde{\vSig} = \vU_M\vS \\ 
& \vS \text{ is diagonal }
\end{array}.  \nonumber
\end{equation}
Note that in this formulation, both $\tilde{\vSig}$ and $\vS$ are diagonal matrices. This, along with the local optimality condition  $\vY'\vSig_x^2\vY\vU_M \tilde{\vSig} = \vU_M \vS$ and the orthonormality condition $\vU_M'\vU_M = \vI_m$, implies that the columns of $\vU_M$ can be chosen as the eigenvectors of  $\vY'\vSig_x^2\vY$. At these locally optimal points we have 
\begin{align}
\vS = \mathbf{\Lambda}\tilde{\vSig}, \nonumber 
\end{align}
where  $\mathbf{\Lambda}$ is the diagonal matrix containing $m< n$ of the eigenvalues of $\vY'\vSig_x^2\vY$ , and the value of the objective function evaluated at these locally optimal $\vU_M$'s is 
\begin{equation}
 \tr\left(\mathbf{\Lambda}\tilde{\vSig}\right) = \tr\left(\vS\right). \nonumber
\end{equation}
If we choose $i^{th}$ column of $\vU_M$ as the $j^{th}$ eigen vector of $\vY'\vSig_x^2\vY$ then the $i^{th}$ diagonal entry of $\vS$ is equal to the product of $j^{th}$ eigenvalue $\vY'\vSig_x^2\vY$ with the $i^{th}$ diagonal entry of $\tilde{\vSig}$. The problem of solving \textbf{P1} is effectively to converted to choosing $m$ eigen vectors out of $n$ eigen vectors of $\vY'\vSig_x^2\vY$ so that the $i^{th}$ largest eigenvalue of $\vY' \vSigma_x^2 \vY$ is multiplied with $i^{th}$ largest value in the set $ \left\{\frac{\sigma_k^2}{1+\sigma_k^2}\right\}_{k=1}^m$. This gives us the optimal solution to \textbf{P1}.

\subsubsection{Final Algorithm}
Based on the analysis in subsections \ref{subsection: Pa} and \ref{subsection: Pb} we propose the algorithm in Table \ref{tab:alg1} to solve problem \eqref{eqn:opt2}. The algorithm  iterates over $\mathbf{P_0}$ and $\mathbf{P_1}$ a total of $N$ times, and the final solution is given by $\vA' = \vY\vU_M^N\vSigma_M^N$. 

\section{Evaluation}\label{sec:eval}
We evaluate the performance of our proposed sensing matrix design procedure via experimentation on synthetic data.  We consider signals of dimension $n=100$, for which the number of signal and clutter models are $m_x = m_c = 10$, and where each model (in each class) is a covariance matrix of rank $6$. The actual covariance matrices of the signal and clutter models are constructed randomly using a (different) random set of $n$ orthonormal $n$-dimensional vectors, and randomly generated (positive) singular values.  

For a subset of possible values of $m$ we perform $1000$ trials of the following experiment.  First, we select one model randomly from the set $\{\vSig_{x,i}\}_{i=1}^{m_x}$ and generate $\vx$ as a zero-mean Gaussian random vector having this covariance matrix, and we generate $\vc$ similarly using one model selected randomly from $\{\vSig_{c,i}\}_{i=1}^{m_c}$.  We then generate four different sets of observations $\vy_i$ obtained using corresponding measurement matrices $\vA_i$, for $i=1,\dots,4$, as follows.  First, For $\vC$ an $m\times n$ matrix whose elements are iid $\mathcal{N}(0,1/m)$ random variables, we let $\vA_1 = \vC(\alpha \vI_n)$ denote observations obtained by traditional random projections. Next, we let $\vA_2 =  \vA^*$, where $\vA^*$ is the solution of \eqref{eqn:opt2} corresponds to the sensing matrix designed via our approach.  

We also compare with two more ``heuristic'' approaches -- in the first of these, we form the sensing matrix $\vA_3$ from a low rank approximation of the Wiener filter for estimating $\vx$ from the mixture $\vx + \vc$, as discussed in \cite{scharf}.  Specifically, here we form $\vW_{lr} = \vB (\vSig_x + \vSig_c)^{-1/2}$ where $\vB$ is the best rank $m$ approximation of $\vSig_x(\vSig_x + \vSig_c)^{-1/2}$ (in the least-squares sense).  Then, we let $\vW_{lr} = \vU_{\vW_{lr}}  \vSig_{\vW_{lr}}\vV_{\vW_{lr}}' $ and obtain the sensing matrix $\vA_3$  by retaining the first $m$ rows of the matrix $ \vSig_{\vW_{lr}}\vV_{\vW_{lr}}'$, and appropriately rescaling to meet the sensing energy constraint. This represents a case where we employ a classic (linear) estimation strategy directly into the measurement process while keeping in mind that we are allowed only to take $m$ measurements.  We also investigate the estimation performance associated with the sensing matrix $\vA_4 = \check{\vA}^*$, where $\check{\vA}^*$ represents the solution of \eqref{eqn:opt2} in a modified setting where clutter models are not viewed as clutter, but rather as additional \emph{signal} models.  In words, this describes the case where we design $\vA$ in order to accurately estimate the mixture $\vx + \vc$, deferring the separation entirely to a subsequent step. The additive noise in each case is $\vw\sim\mathcal{N}(0, \vI_{m\times m})$. 

We aim to reconstruct the signal $\vx$ in each case using a group lasso approach \cite{yuan} that explicitly leverages the correlation structure described by each model.  To that end, we let $\vD_x$ be the $n \times (6\cdot m_x)$ signal dictionary whose $n\times 6$ blocks correspond to the top $6$ eigenvectors of each of the signal models $\vSigma_{x,1}, \dots, \vSigma_{x,m_x}$, and similarly for $\vD_c$, and we denote by $\vD = [\vD_x~\vD_c]$ the combined dictionary, comprised of a total of $m_x + m_c$ models. Then, we obtain the estimates for each sensing matrix/observation vector pair as $\widehat{\vx}_i = [\vD_x \ \v0] \ \left [\arg \min_{\vbeta} \|\vy_i - \vA_i \vD \vbeta\|^2_2 + \lambda \Omega(\beta) \right]$, for $i=1,\dots,4$ where the parameter vector $\vbeta$ is $6\cdot (m_x+m_c)\times 1$ and the regularizer $\Omega(\beta) = \sum_{j = 1}^{m_x + m_c} {\sqrt{\mathbf{v}_j'\vLambda^{-1}_j \mathbf{v}_j}}$, where  $ \mathbf{v}_j = \beta_{[6(j-1) + 1: 6j]}$ is a sub vector of the parameter vector corresponding to the $j$-th overall model and $\vLambda_j$ is the $6\times 6$ diagonal matrix whose elements are the nonzero eigenvalues of the $j$-th model.  Optimizations were performed using the Sparse Modeling Software (SpaMS) \footnote{Available online at {\tt http://spams-devel.gforge.inria.fr}}.
 
We compare the performance of each of the four approaches in terms of reconstruction SNR vs. sensing energy budget $\alpha^2$.  The results, depicted in Figure~\ref{fig:res}, show that our proposed approach (blue line, circle marker) outperforms each of the other approaches -- traditional CS (black dotted line, square markers), the sensing approach based on the low rank Wiener filter (green dotted line, circle markers), and the approach where the clutter models are viewed as signal models and separation is left to the final estimation step (red dotted line, triangle markers) -- across all sensing energy budgets, and for each subsampling case examined ($m=20, 40, 60, 80$ measurements, respectively, in panels (a)-(d)).

\section{Conclusions}\label{sec:conc}
It is interesting to see that both our proposed approach as well as the low rank Wiener filter approach are performing a kind of ``annihilate-then-estimate'' sensing strategy, while the approach corresponding to the sensing matrix $\vA_4$ is more of an ``estimate-then-annihilate'' strategy.  Our results here suggest that the former approach is more viable here -- in other words, our empirical results here suggest that we should incorporate some ``cancellation'' into the sensing matrix itself, rather than relying on the final estimation step to perform the separation.  Of course, these observations are based on a preliminary investigation, and further investigation is needed in order to make such claims definitively.

Further, while our design approach was based on a MSE minimization criteria, we note a point of comparison between our approach and related design strategies that are based on maximizing mutual information between the vector $\vx$ to be estimated and observations obtained for a specific $\vA$.  At first glance, these criteria are (seemingly) different, however, a fundamental connection between the minimum MSE matrix and the mutual information between the unknown $\vx$ and the observations $\vy$ (more specifically, its gradient with respect to various problem parameters, such as the matrix $\vA$) has recently been established -- see \cite{Palomar}.  Indeed, the work \cite{Palomar} discusses a related task of linear pre-coder design in an effectively ``clutter-free'' scenario, and proposes a gradient projection approach for obtaining the optimal precoder matrix.  A more thorough investigation of the similarities between this work and the method we examine here is deferred to a subsequent effort.  

\section*{Acknowledgments}
The authors thank Nikhil Rao and Rob Nowak for enlightening discussions, and for collaborating on an initial investigation into this problem \cite{SPARS}.
\bibliographystyle{IEEEbib}
\bibliography{clutterbib}
\end{document}